\pdfoutput=1

\documentclass[11pt]{article}

\usepackage{authblk}
\usepackage{acl}

\usepackage{times}
\usepackage{latexsym}
\usepackage{amsmath}
\usepackage{amsfonts}
\usepackage{caption}
\usepackage{subcaption}
\usepackage{graphicx}
\usepackage{booktabs}
\usepackage{hyperref}
\usepackage{float}

\usepackage{footnote}
\makesavenoteenv{tabular}
\makesavenoteenv{table}

\usepackage[T1]{fontenc}

\usepackage[utf8]{inputenc}

\usepackage{microtype}
\usepackage{bm}

\usepackage{amstext} 
\usepackage{array}   
\newcolumntype{C}{>{$}c<{$}} 

\newcommand{\B}{\bm}
\newcommand{\U}{\underline}
\newcommand{\code}{\texttt}

\newcommand{\Attn}{\text{Attn}}
\newcommand{\cat}{\text{cat}}

\usepackage[symbol]{footmisc}

\title{Prefix-Propagation: Parameter-Efficient Tuning for Long Sequences}

\author[2*]{Jonathan Li}
\author[1,2]{Will Aitken}
\author[1,2]{Rohan Bhambhoria}
\author[1,2\textsuperscript{\textdagger}]{Xiaodan Zhu}
\affil[1]{\hspace{1pt}Department of Electrical and Computer Engineering, Queen’s University}
\affil[2]{\hspace{1pt}Ingenuity Labs Research Institute, Queen’s University}

\affil[ ]{\texttt{\{jxl, will.aitken, r.bhambhoria, xiaodan.zhu\}@queensu.ca}}

\begin{document}
\maketitle
\begingroup\def\thefootnote{*}\footnotetext{Work done during a student internship at Ingenuity Labs.}\endgroup
\begingroup\def\thefootnote{\textdagger}\footnotetext{Corresponding author.}\endgroup
\renewcommand{\thefootnote}{\arabic{footnote}}
\begin{abstract}
Parameter-efficient tuning aims to mitigate the large memory requirements of adapting pretrained language models for downstream tasks. For example, one popular method, prefix-tuning~\citep{li-liang-2021-prefix,liu-etal-2022-p}, prepends trainable tokens to sequences while freezing the rest of the model's parameters. Although such models attain comparable performance with fine-tuning when applied to sequences with short to moderate lengths, we show their inferior performance when modelling long sequences. To bridge this gap, we propose \textit{prefix-propagation}, a simple but effective approach that conditions prefixes on previous hidden states. We empirically demonstrate that prefix-propagation outperforms prefix-tuning across long-document tasks, while using $\sim$50\% fewer parameters. To further investigate the proposed architecture, we also show its advantage in calibration, and perform additional study on its relationship with kernel attention. To the best of our knowledge, this work is the first to focus on parameter-efficient learning for long-sequence language tasks.\footnote{Our code is publicly available at \url{https://github.com/MonliH/prefix-propagation}}
\end{abstract}

\section{Introduction}
The Transformer architecture \citep{vaswani-et-al-2017-attention} has changed the landscape of recent natural language processing approaches by enabling the pre-training of state-of-the-art large language models (LLM) \citep{devlin-etal-2019-bert, he2020deberta, Brown:20}. However, fine-tuning and storing full copies of LLMs can consume prohibitively large quantities of resources. Parameter-efficient fine-tuning (PEFT) methods such as prefix-tuning \citep{li-liang-2021-prefix, He:21, liu-etal-2022-p} address these concerns by reducing the number of trainable parameters. Prefix-tuning can tune 0.01\% of parameters and still match the performance of regular fine-tuning (updating all model parameters). 

\begin{table}
\centering
\small
\resizebox{0.9\columnwidth}{!}{%
\begin{tabular}{lccc}
\toprule
Method & 20-newsgroups & Hyperpartisan \\
\midrule
Prefix-Tuning  & 69.7 & 75.3 \\
Fine-Tuning  & 72.3 & 81.5 \\
\bottomrule
\end{tabular}}
\caption{Mean F1-Scores of prefix-tuning and fine-tuning \code{Longformer} for common long-document classification tasks.}
\label{tab:prefix_is_bad}
\end{table}

PEFT has been investigated for tasks with inputs consisting of sentences, sentence-pair, or sequences that fit within the typical LLM maximum tokens.  
However, the performance of PEFT for tasks with longer textual sequences has been overlooked. In this work, we investigate this oversight and provide evidence suggesting that the gap between PEFT and regular fine-tuning is substantial when modelling long sequences.
As shown in Table \ref{tab:prefix_is_bad}, prefix-tuning underperforms fine-tuning on long sequence classification tasks, Hyperpartisan \citep{kiesel-etal-2019-semeval} and 20-newsgroups \citep{Lang95}, when used with the popular long-document model \code{Longformer} \citep{Beltagy:20}.

\begin{figure*}[t]
    \centering
    \begin{subfigure}[t]{0.46\textwidth}
        \includegraphics[width=\textwidth]{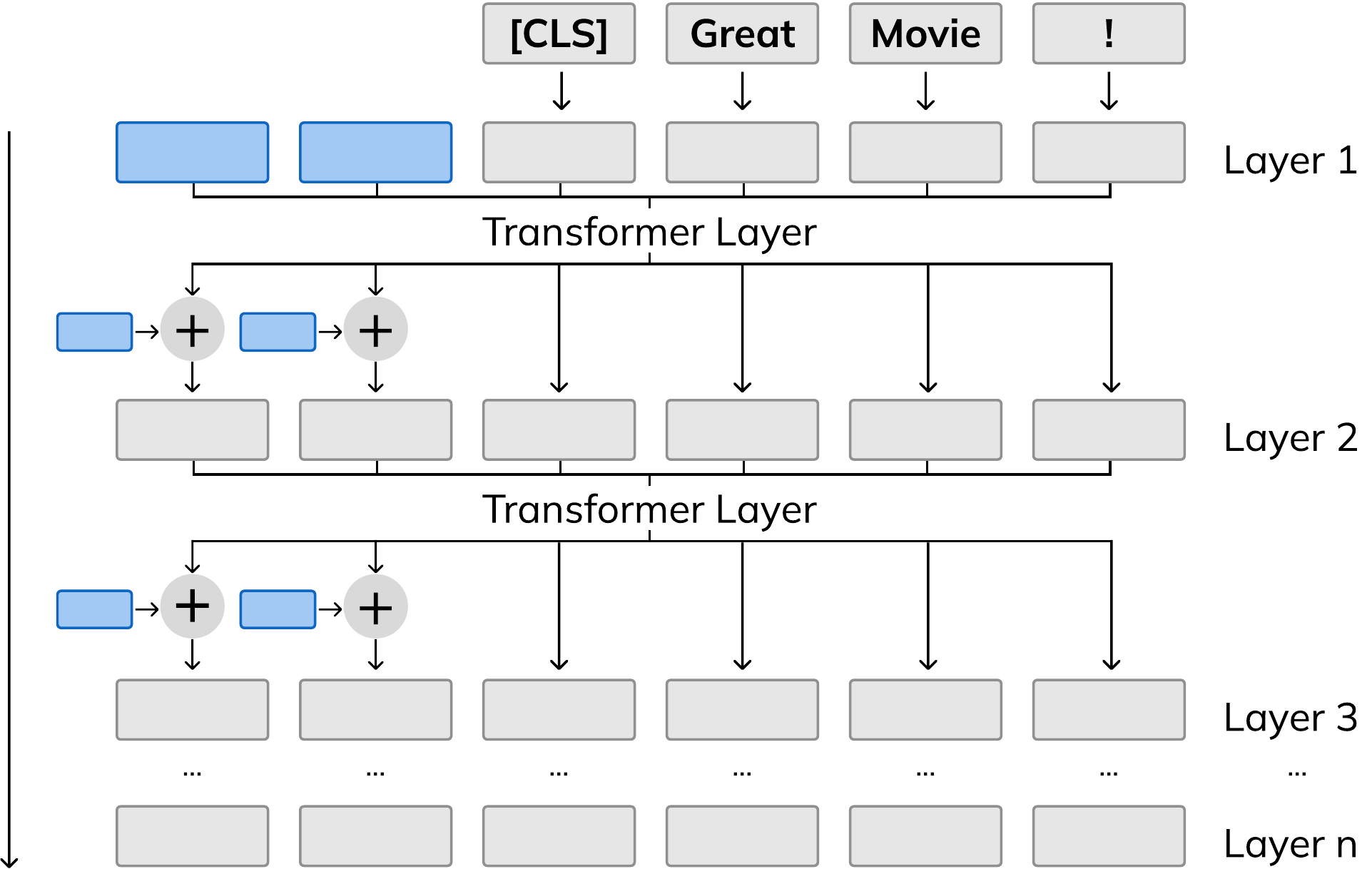}
        \caption{Prefix-Propagation}
        \label{fig:prefixPropagation}
    \end{subfigure}
    \hfill
    \begin{subfigure}[t]{0.5\textwidth}
        \includegraphics[width=\textwidth]{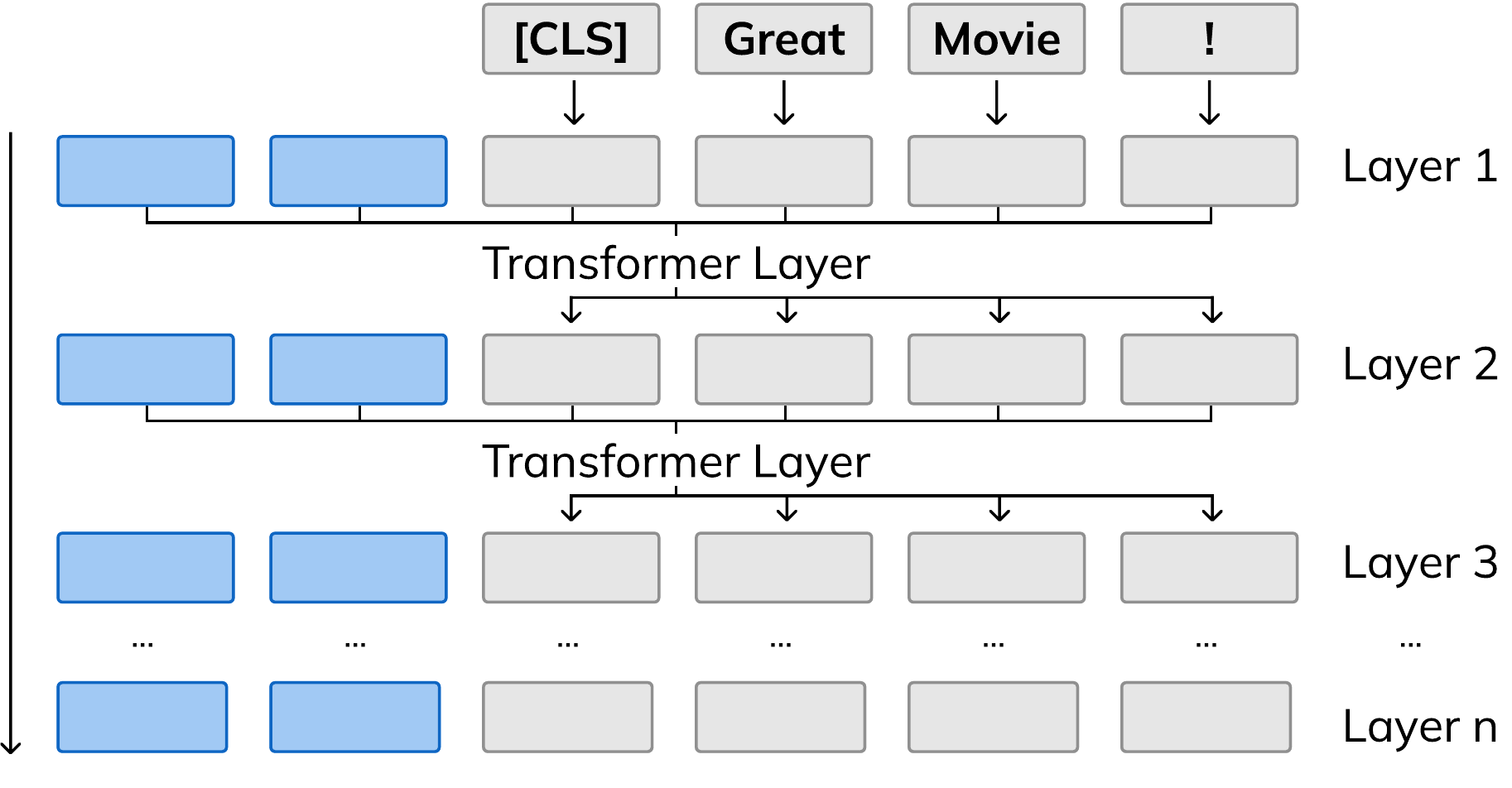}
        \caption{Prefix-Tuning}
        \label{fig:prefixTuning}
    \end{subfigure}
    \caption{Illustration of the differences between (\subref{fig:prefixPropagation}) prefix-propagation (ours) (\subref{fig:prefixTuning}) and prefix-tuning \citep{liu-etal-2022-p,li-liang-2021-prefix}. Blue blocks denote trainable prompts, and ``Transformer Layer'' represents the computation done in a layer of the pre-trained LLM. Note that in prefix-propagation (\subref{fig:prefixPropagation}), the summation of prefixes continues for layers beyond 3, up to $n$. This operation is encapsulated by the ellipses. In prefix-tuning (\subref{fig:prefixTuning}), prefixes in subsequent layers do not depend on hidden states from past layers (they are simply overwritten).}
    \label{fig:prefixVsPropagation}
\end{figure*}

In this paper, we propose a simple and effective method, \textit{prefix-propagation}, which consistently improves the performance of PEFT for long sequence models. Unlike prefix-tuning, prefix-propagation propagates the hidden states corresponding to prefixes through the attention computation. This allows for the prefixes hidden states to dynamically change as the input propagates through each layer. To further understand prefix propagation, we investigate the reliability of the model's predictions by performing analyses on calibration. Lastly, we conduct study on prefix-based methods in terms of kernel attention to strengthen their theoretical value.

In summary, our contributions are as follows:

\begin{itemize}
    \item  We study PEFT for long documents and show that prefix-tuning is significantly inferior to fine-tuning in this scenario. To the best of our knowledge, this is the first work to focus on PEFT for long documents.

    \item We introduce prefix-propagation, which consistently improves the performance over prefix turning on the different long document datasets, while using 50\% fewer parameters. 

    \item We study the reliability of the predictions by performing analyses on calibration and show that models tuned with prefix-propagation are better calibrated. 
    
    \item We elucidate the relationship between prefix-propagation and kernel attention and perform an ablation study that utilizes this insight. 
    
\end{itemize}

\section{Related Works} 
\paragraph{Long Sequence Models} Numerous methods have been proposed to reduce the complexity of attention from $O(n^2)$ to $O(n)$ such as kernel approximations \citep{Choromanski:20, Katharopoulos:20, Peng:21} and fixed \citep{Child:19, Beltagy:20, Zaheer:20} or learned \citep{Kitaev:20} sparse attention patterns. For a broader summary, please refer to \citet{Tay:22}. In this work, we use \code{Longformer} \citep{Beltagy:20}. To linearize attention complexity, \code{Longformer} employs sliding window attention while globally attending to relatively few special tokens. 

\paragraph{Parameter-Efficient Tuning}  Inspired by the success of manual prompting \citep{Brown:20}, prefix-tuning \citep{li-liang-2021-prefix,liu-etal-2022-p} prepends trainable ``soft'' prompts to an input sequence. Although further PEFT methods have since been introduced \citep{He:21, Hu:21, ben-zaken-etal-2022-bitfit}, we focus on adapting prefix-tuning. We note that our adaptation does not violate orthogonality and thus prefix-propagation can still be compounded with other PEFT methods as proposed in the UnifiedPET framework \citep{He:21}, likely yielding similar performance gains. We leave the empirical validation of this hypothesis for future work.

Out work also adheres to the key motivation of the recent PEFT method, inducer-tuning \cite{Chen:22}, which is that optimal prefixes should be close to queries within their latent space. We derive queries, keys, and values from the same prefix token, limiting the distance that separates them. 

\begin{table*}[t]
    \centering
    \renewcommand\arraystretch{1.3}
    \resizebox{\textwidth}{!}{%
    \begin{tabular}{l|c|CCCCCCCCCCCC}
        \toprule
         \multicolumn{1}{c|}{Method} & \% Tuned & \text{WikiHop} & \multicolumn{3}{c}{ArXiv} & \multicolumn{3}{c}{20-newsgroups} & \multicolumn{3}{c}{Hyperpartisan} \\
         & & Acc & P & R & F1 & P & R & F1 & P & R & F1 \\
         \midrule
         \code{RoBERTa} PT         & 0.1  & 11.7 & 79.4 & 79.6 & 79.8 & 67.9 & 67.0 & 68.2 & 70.4 & 59.2 & 64.1 \\
         Prefix-Tuning      & 0.1  & 38.9       & 81.5 & 81.7 & 82.7 & 68.9 & 68.4 & 69.7 & 78.3 & 73.8 & 75.3 \\ 
         Prefix-Propagation & 0.05 & \U{42.2}   & \B{83.1} & \B{83.1} & \B{83.3} & \U{70.1} & \U{69.7}  & \U{71.0} & \U{86.4} & \B{77.7} & \B{81.8} \\
         Fine-Tuning        & 100  & \B{74.0}   & \B{83.1} & \U{82.9} & \B{83.3} & \B{71.8} & \B{71.2} & \B{72.3} & \B{87.8} & \U{76.2} & \U{81.5} \\
         \bottomrule
    \end{tabular}}
    \caption{Main results of prefix-propagation compared to prefix-tuning and traditional fine-tuning on the validation sets of each dataset. All approaches use \code{Longformer-base} except ``\code{RoBERTa} PT'', which is prefix-tuning on \code{RoBERTa-base}. Micro F1 and macro-average precision (``P'') and recall (``R'') is reported for ArXiv, Hyperpartisan (with mean across 5 runs), and 20-newsgroups. Accuracy is reported for WikiHop. Performance is reported on test splits with the exception of Hyperpartisan, which is performance on the validation split (See Appendix \ref{sec:experimentDetails} for reasoning). The best run is \textbf{bold} and second best is \underline{underlined}.}
    \label{tab:mainResults}
\end{table*}
\setlength{\tabcolsep}{6pt}

\section{Prefix Propagation}
\subsection{Methodology}
\label{sec:prefixPropOutline}
In this section we introduce prefix-propagation, which, unlike prefix-tuning, propagates the hidden states corresponding to prefixes through the attention computation. This allows for the prefixes hidden states to dynamically change as the input propagates through each layer. Prefix-propagation and its predecessor, prefix-tuning are depicted in Figure \ref{fig:prefixPropagation} and \ref{fig:prefixTuning} respectively. For the first layer of the transformer, we prepend $j$ trainable prefixes (i.e., embeddings) to the input sequence (blue blocks in top left of Figure~\ref{fig:prefixPropagation}). Then, before every subsequent layer, we sum new trainable matrices onto the first $j$ embeddings corresponding to the prefixes (denoted by the sum operators in Figure~\ref{fig:prefixPropagation}). By propagating instead of overwriting, we halve the number of parameters trained while simultaneously improving performance on long-document tasks.

We now formalize prefix-propagation. Multi-headed attention processes query, key, and value matrices derived from a sequence $C \in \mathbb{R}^{m\times d}$ with length $m$ and embeddings of size $d$. Our method modifies traditional attention by concatenating a prefix $P\in \mathbb{R}^{j\times d}$ of length $j$ to the sequence:
\begin{align}
H_{l,i} &= \Attn (\begin{aligned}[t]&D^{(l)}W_q^{(l,i)}, \\
&D^{(l)}W_k^{(l,i)}, D^{(l)}W_v^{(l,i)})\end{aligned} \label{eqn:prefixPropagation} \\
D^{(l)} &=
\begin{cases}
    \cat (P^{(l)}, C) & \text{if $l=1$} \\
    \cat (P^{(l)}+C[{:}j, :], C[j{:}, :]) & \text{if $l>1$} \\
  \end{cases}\nonumber
\end{align}
where inputs $C$ are projected through pre-trained weight matrices $W_q^{(l, i)}, W_k^{(l, i)}, W_v^{(l, i)}\in \mathbb{R}^{d\times d_h}$ per layer $l$ and head $i$ yielding the output of the attention head, $H \in \mathbb{R}^{(j+m)\times d_h}$. The prefixes are concatenated for the first layer ($l=1$) and summed to their corresponding hidden states for the remaining layers ($l>1$). We do not continually concatenate new prefixes to the sequence to avoid increasing the sequence length after each layer.

For both prefix-tuning and prefix-propagation, prefixes (keys and values) are globally attended to by all queries. Unlike prefix-tuning however, our method concatenates additional hidden states \textit{before} the hidden states $C$ are projected by $W_k^{(i)}$ and $W_v^{(i)}$. By doing so, prefix-propagation modifies query matrices, allowing prefixes to attend to other hidden states globally, thereby increasing representation capability. This approach is somewhat analogous to the external global tokens inserted in the \code{BigBird-ETC} model \citep{Zaheer:20}. By attending to other tokens, the prefixes can act as special storage tokens, which is particularly useful in the restricted regime of long-document modelling where relatively few tokens have global context. Conversely, prefix-tuning only concatenates trained key and value matrices, $P_k, P_v \in \mathbb{R}^{j\times d_h}$, statically to the sequence:
\begin{align}
H_{l,i} = \Attn (&CW_q^{(l, i)}, \cat (P_k^{(l, i)}, CW_k^{(l, i)}), \nonumber \\
             &\cat (P_v^{(l, i)}, CW_v^{(l, i)})) \label{eqn:prefixTuningAttn}
\end{align}

 Since our method has a single prefix matrix, $P$ instead of separate $P_k$ and $P_v$ matrices, we reduce the number of trained parameters by 50\%.

\subsection{Calibration}
We further study the proposed prefix-propagation method to understand the reliability of model's predictions through calibration. Well-calibrated models output confidence scores that closely match the models' accuracy. Either over-confident or under-confident models are undesirable. 
Calibration has widely been overlooked in PEFT methods. To quantify calibration in our work, we use expected calibration error (ECE), which bins predictions based on model confidence and compares them to accuracy \citep{naeini-etal-2015-obtaining,guo-etal-2017-on}.

\subsection{Kernel Decomposition}
Traditional attention is analogous to applying a kernel smoother over inputs \citep{Tsai:19}. Motivated by this insight, we reformulate prefix-propagation as a sum of kernelized attention modules. Separating the modules introduces flexibility in two ways: (1) Their individual kernel forms can be mixed and matched and (2) A hyperparameter scale factor $\alpha$ can be applied to the prefix component to increase or decrease its \mbox{weighting}. Equation \ref{eqn:prop_kern} defines kernel decomposition for prefix-propagation\footnote{We omit layer, $l$ and head, $i$ for brevity.}:
\begin{align}
H = \;&\text{Kern}(\cat (P, C)W_q,CW_k,CW_v) \nonumber\\
+\;(\alpha)&\text{Kern}(\cat (P, C)W_q,PW_k,PW_v) \label{eqn:prop_kern}
\end{align}
where Kern refers to kernel attention as formulated in \citep{Tsai:19}. The first term results from attending to the original sequence, $C$, and the second comes from attending to the prefixes, $P$. We provide the derivation of Equation \ref{eqn:prop_kern} and the full definition of kernel attention in Appendix \ref{sec:prefixRewrite}. 

Our main motivation for presenting prefix decomposition is to establish foundational knowledge and guide future research. Ergo, we restrict experiments in this initial presentation to using just the default exponential kernel (Appendix \ref{sec:prefixRewrite}).

\section{Experiments and Results}

\paragraph{Datasets} We evaluate our approach on three long-document classification tasks: ArXiv \citep{he-arxiv}, an 11-class classification task composed of academic research papers, the 20-newsgroups \citep{Lang95} classification task consisting of mailing lists that fall into one of 20 classes, and the Hyperpartisan dataset, a binary classification task for extremist news classification \citep{kiesel-etal-2019-semeval}. We also run experiments on WikiHop \citep{welbl-etal-2018-constructing}, a long-document reading comprehension task requiring multi-step reasoning. 

Due to compute limitations inherent to working with long documents, with the exception of Hyperpartisan, we only report a single run for each task. This mimics the original \code{Longformer} reporting scheme \citep{Beltagy:20}. For Hyperpartisan, the smallest of the datasets, we report mean metrics averaged over five seeds. 

\paragraph{Baselines} As a baseline, we fine-tune \code{Longformer-base} (approx. 149M parameters) as closely as possible to \citet{Beltagy:20}. For PEFT, we evaluate prefix-tuning on \code{Longformer-base} and \code{RoBERTa-base} (approx. 125M parameters) \citep{liu-et-al-2019-roberta}. 

More details on dataset sizes, pre-processing, and hyperparameters are in Appendix~\ref{sec:experimentDetails}.

\subsection{Results and Discussion}

\begin{table}[t]
    \centering
    \renewcommand\arraystretch{1.3}
    \resizebox{\linewidth}{!}{%
    \begin{tabular}{l|CCC}
        \toprule
        \multicolumn{1}{c|}{Method} & \multicolumn{1}{c}{ArXiv} & \multicolumn{1}{c}{HY.} & \multicolumn{1}{c}{NG.} \\
         \midrule
         \code{RoBERTa} PT  & \U{0.056} & 0.228 & 0.123 \\
         Prefix-Tuning      & 0.075 & 0.153 & \B{0.117} \\
         Prefix-Propagation & \B{0.042} & \B{0.093} & \U{0.122} \\
         Fine-Tuning        & 0.099 & \U{0.138} & 0.212 \\
         \bottomrule
    \end{tabular}}
    \caption{ECE scores of tested approaches. Lower is better. \textbf{Bold} is the best and \underline{underline} is the second best. ``HY.'' is Hyperpartisan, and ``NG.'' is 20-newsgroups.}
    \label{tab:calibration}
\end{table}

Across all tasks, our results in Table \ref{tab:mainResults} verify that prefix-tuning is inferior to fine-tuning long sequences. Conversely, prefix-propagation consistently outperforms prefix-tuning and is comparable to fine-tuning on most tasks. Prefix propagation also performs competitively on Hyperpartisan, a relatively small dataset with only 625 samples. This is in contrast to prefix-tuning, which is known to underperform in low-data settings \citep{gu-etal-2022-ppt}. Because we ran multiple seeds on Hyperpartisan, we also found that prefix-propagation's better performance relative to prefix-tuning is statistically significant ($p<0.05$, using a single-tailed t-test). We do not have multiple samples to run these tests for larger datasets, but we emphasize that Hyperpartisan likely has the most variance and yet it is still statistically significant. We suspect that prefix-propagation's performance exceeds prefix-tuning because propagated prefixes can transmit global context across multiple layers, possibly modelling more expressive abstractions. 

\begin{figure}
    \centering
    \resizebox{\linewidth}{!}{
    \includegraphics{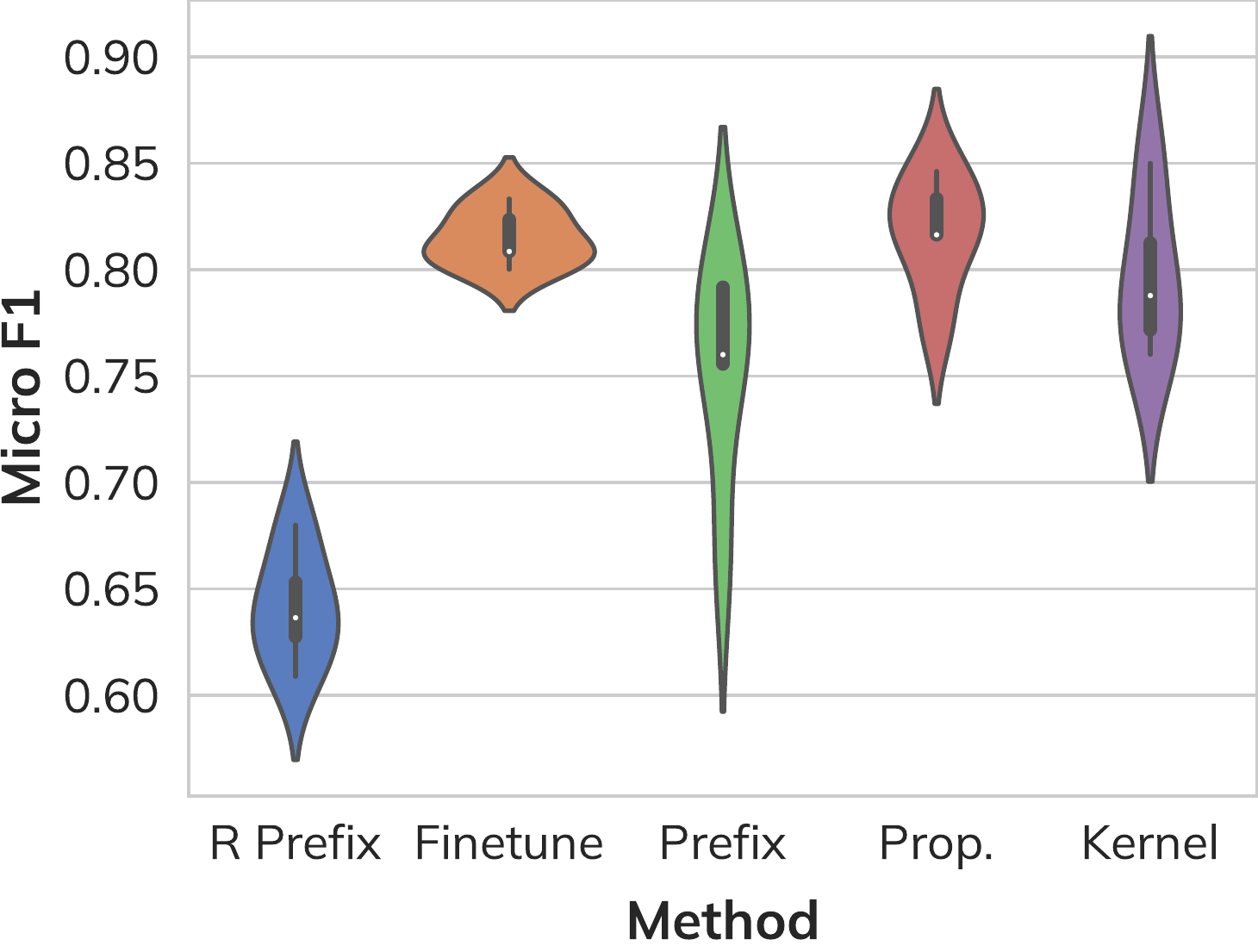}
    }
    \caption{Violin plot of Micro F1 Score for five different seeds on the Hyperpartisan task. White dots, gray boxes, and gray lines are the medians, interquartile ranges, and ranges respectively. Width of the five violin shapes show the probability densities for the corresponding F1-score. All methods tune \code{Longformer-base} except ``R Prefix'', which is prefix-tuning on \code{RoBERTa-base}.}
    \label{fig:hyperpartisanAll}
\end{figure}

We note one exception where prefix-based methods still leave room for improvement: multiple-choice question answering on WikiHop. We hypothesize that prefix methods have insufficient capacity to properly model complex long-document multi-step question answering.

We also observe that prefix-based methods, and especially prefix-propagation, achieve better calibration than fine-tuning, as shown in Table~\ref{tab:calibration}. Unlike prefix-tuning however, prefix-propagation effectively balances calibration with accuracy metrics. The calibration of fine-tuning deteriorates as training progresses (Figure \ref{fig:arxivCalibration} in Appendix~\ref{sec:ece_app}) and we speculate that this may be due to catastrophic forgetting \citep{https://doi.org/10.48550/arxiv.2207.00099}. 

As an initial test for our ongoing prefix-propagation kernel study, we show results on Hyperpartisan in Figure~\ref{fig:hyperpartisanAll}. The kernelized version of prefix-propagation achieves the best single-run performance, but has higher variance than fine-tuning and prefix-propagation which necessitates further research. 

\section{Conclusion}
Our research focuses on parameter efficient tuning for long documents tasks. We introduce prefix-propagation, which consistently improves performance over prefix-turning on long document datasets, while using 50\% fewer parameters. 
We study the reliability of the predictions by performing analyses on calibration and show that models tuned with prefix-propagation are better calibrated. 
We lastly explicate prefix-propagation from a kernel perspective, uncovering insights for future PEFT research.
\section*{Limitations}

\subsection*{Scope}
This short paper serves as an initial step toward PEFT for long-document models. As such, our evaluated scope of models, tasks, datasets, and kernel variations is limited. We acknowledge the need to experiment across broader settings and hope our work provides a foundation for others to build on. 

Future experiments should analyze the validity and efficacy of using prefix-propagation with other long-sequence models to determine whether the prefix modality is suitable for non-sparse attention approximations. For example, would the projection of prefix vectors using a random feature map as in \citet{Choromanski:20} result in an excessive loss of information for these critical tokens? 

Regarding tasks and datasets, the performance degradation in prefix methods for WikiHop deserves significant attention. Verifying whether this extends to other reading comprehension and question-answering tasks will assist in guiding future research efforts. We restricted our research to the encoder-only version of \code{Longformer}, but using the encoder-decoder version, \code{LED} would enable analysis of sequence-to-sequence tasks. The SCROLLS benchmark \citep{shaham2022scrolls} would be a good starting point for this analysis since it includes an \code{LED} baseline. 

Combining prefix and kernel methods is an ongoing research effort and there are several questions we plan to address: (1) What are the effects of swapping the default exponential kernel with other variants such as linear, polynomial, and RBF? (2) Does making the $\alpha$ scale parameter trainable improve performance? (3) Can we have a separate scale parameter for each query and should they be trainable? (4) Is this approach effective for modalities other than long-document? (5) Can we separate other components of attention into modular kernels (e.g. local and global kernels for sparse attention)?

\subsection*{Robustness}
The size and nature of long-sequence tasks often resulted in long run times for the larger datasets ArXiv, 20-newsgroup and WikiHop. Consequently, we report results of one seed after doing a hyperparameter search for learning rate. This aligns with the reporting system of the original \code{Longformer} paper \citep{Beltagy:20} but greater assurance in all long-sequence task performance could be achieved by accumulating results over several seeds. The size of datasets and iteration over several epochs somewhat mitigate this concern. 

\section*{Ethics Statement}
Our work helps to address the environmental and equitable distribution concerns of LLMs \citep{Strubell:19}. All PEFT variants attempt to reduce resource requirements, primarily via GPU memory consumption and storage requirements. By applying prefix-tuning and our variation, prefix-propagation to long-document models we limit carbon emissions and increase accessibility for low-resource groups. We note that prefix-propagation neither exacerbates nor alleviates other ethical risks such as biases regarding gender, race, religion, etc. that are often embedded in pre-trained LLMs. If such biases exist in the pre-trained model, they will be propagated to downstream tasks regardless of tuning method.

\section*{Acknowledgements}
This research is supported by NSERC Discovery Grants. The second author is also supported by the Vector Scholarship in Artificial Intelligence.

\bibliography{anthology,custom}

\appendix

\section{Kernel Decomposition Derivation}
\label{sec:prefixRewrite}

In the unified framework of \citet{he-etal-2021-towards}, we can write the first layer $l=1$ attention mechanism of prefix-propagation as:

\begin{align} \label{eqn:app_orig}
H_{l,i} = \Attn (&\cat (P^{(l)}, C)W_q^{(l)(i)}, \\
&\cat (P^{(l)}, C)W_k^{(l)(i)}, \nonumber \\
&\cat (P^{(l)}, C)W_v^{(l)(i)}) \nonumber 
\end{align}
where $P$ is a trained prefix for each downstream task. Omitting layer and head indices and using $D=\cat(P,C)$ for brevity, we can rewrite Equation \ref{eqn:app_orig} as:
\begin{align} \label{eqn:app_uni_pet}
\begin{split}
H & = \Attn (DW_q,\cat (P, C)W_k, \cat (P, C)W_v) \\
& = \text{softmax}(DW_q\cat (PW_k, CW_k)) \begin{bmatrix}PW_v\\CW_v\end{bmatrix} \\
& = \begin{aligned}[t]
(1-&\lambda(C))\text{softmax}(DW_qW_k^{\top}C^{\top})CW_v\\
+ &\lambda(C)\text{softmax}(DW_qW_k^{\top}P^{\top})PW_v
\end{aligned} \\
& = \begin{aligned}[t]
(1-&\lambda(C))\Attn (DW_q,CW_k,CW_v) \\
+&\lambda(C) \Attn (DW_q,PW_k,PW_v) 
\end{aligned} \\
& = \begin{aligned}[t]
(1-&\lambda(C))\Attn (\cat(P,C)W_q,CW_k,CW_v) \\
+&\lambda(C) \Attn (\cat(P,C)W_q,PW_k,PW_v) 
\end{aligned}
\end{split} 
\end{align}
where $\lambda(C)$ is a scalar (dependent on $C$) to normalize softmax over the sequence and the prefixes and is computed by:
\begin{align}
    \lambda(C)&=\frac{\sum_{i} DW_qWk^{\top}P^{\top}}{\sum_{i}DW_qW_k^{\top}P^{\top}+\sum_{j}DW_qW_k^{\top}C^{\top}} 
\end{align}
 We consider the two terms of Equation \ref{eqn:app_uni_pet} as kernelized attention modules which brings us back to the complete kernel decomposition:
\begin{align}
H = \;&\text{Kern}(\cat (P, C)W_q,CW_k,CW_v) \nonumber\\
+\;(\alpha)&\text{Kern}(\cat (P, C)W_q,PW_k,PW_v)
\end{align}
where $\alpha$ is an introduced hyperparameter that replaces the fixed weighting of $\lambda$. This change allows us to explicitly increase the weighting of prefixes by scaling the prefix kernel's coefficients. $\text{Kern}$ is the kernelized attention variant described in \citet{Tsai:19}: 
\begin{equation} \label{eqn:kern_att}
\text{Kern}(Q, K, V)_i 
= \sum_{j=1}^{N}\frac{k(Q_i,K_j)}
        {\sum_{j'=1}^{N}k(Q_i,K_{j'})}V_j
\end{equation} 
where subscripts (e.g. $i$) index the rows of a matrix, $N$ is the number of key and value vectors, and $k$ is a kernel function that calculates the similarity score between two vectors. We do not experiment with altering the kernel type since the default exponential kernel inherent to softmax attention already implicitly maps the input vectors to an infinite feature space. Therefore, the kernel function in Equation \ref{eqn:kern_att} takes the form:
\begin{equation}
k(x_q,x_k) = exp\left(\frac{\langle x_q,x_k \rangle}{\sqrt{d_k}}\right)
\end{equation}
where $\langle \cdot , \cdot \rangle$ signifies the dot product and $d_k$ is the dimension of key projections.

\section{Experimental Details}
\label{sec:experimentDetails}

\begin{table}
\resizebox{\columnwidth}{!}{%
\begin{tabular}{lcc}
\toprule
Artifact & Version & License \\
\midrule
\code{transformers} \citep{Wolf:20}\footnote{\href{https://github.com/huggingface/transformers}{https://github.com/huggingface/transformers}} & 4.23.1 & Apache 2.0 \\  
\code{datasets} \citep{Datasets:21}\footnote{\href{https://github.com/huggingface/datasets}{https://github.com/huggingface/datasets}} & 2.6.1 & Apache 2.0 \\
\code{GPyTorch} \citep{GTorch}\footnote{\href{https://github.com/cornellius-gp/gpytorch}{https://github.com/cornellius-gp/gpytorch}} & 1.9.0 & MIT \\
\code{RoBERTa} \citep{liu-et-al-2019-roberta}\footnote{\href{https://huggingface.co/docs/transformers/model_doc/roberta}{https://huggingface.co/docs/transformers/model\_doc/roberta}} & \code{base} & MIT \\
\code{Longformer} \citep{Beltagy:20}\footnote{\href{https://huggingface.co/docs/transformers/model_doc/longformer}{https://huggingface.co/docs/transformers/model\_doc/longformer}} & \code{base} & Apache 2.0 \\
P-Tuning \citep{liu-etal-2022-p}\footnote{\href{https://github.com/THUDM/P-tuning-v2}{https://github.com/THUDM/P-tuning-v2}} & 2.0 & Apache 2.0 \\
\midrule
ArXiv \citep{he-arxiv}\footnote{\href{https://huggingface.co/datasets/ccdv/arxiv-classification}{https://huggingface.co/datasets/ccdv/arxiv-classification}} & \code{no\_ref} & Unspecified \\
Hyperpartisan \citep{kiesel-etal-2019-semeval}\footnote{\href{https://github.com/zliucr/hyperpartisan-news-detection}{https://github.com/zliucr/hyperpartisan-news-detection}} & 1.0 & CC BY 4.0 \\
20-newsgroup \citep{Lang95}\footnote{\href{https://scikit-learn.org/1.2/datasets/real_world.html\#newsgroups-dataset}{https://scikit-learn.org/1.2/datasets/real\_world.html}} & 1.0 & Unspecified \\
WikiHop \citep{welbl-etal-2018-constructing}\footnote{\href{http://qangaroo.cs.ucl.ac.uk/}{http://qangaroo.cs.ucl.ac.uk/}} & 1.1 & CC BY SA 3.0 \\
\bottomrule
\end{tabular}}
\caption{Complete list of artifacts used in our experiments along with their versions and licenses.}
\label{tab:artifacts}
\end{table}

\paragraph{Artifact Notes} Table \ref{tab:artifacts} summarizes the complete list of artifacts we used in our experiments along with their licenses and versions. All libraries were used for their intended purpose of open-source development. The ArXiv, Hyperpartisan, and WikiHop datasets were released in research contexts to evaluate and/or develop state-of-the-art algorithms. The intended use of 20-newsgroups is not explicit, although it is commonly used for natural language processing in research. We therefore believe we have adhered to the intended usages of the datasets we included. 

We do not anonymize the data for 20-newsgroups as (a) the trained models is not being deployed (only used for evaluation purposes) and (b) the non-anonymized variant is already publicly available. We chose to use the datasets in the current form for fair comparison with other baselines and therefore did not do a detailed analysis for those artifacts. We refer readers to the cited original works in Table \ref{tab:artifacts} for complete documentation.

\begin{table}
\centering
\resizebox{0.8\columnwidth}{!}{%
\begin{tabular}{c|cccc}
\toprule
 & HY. & ArXiv & NG. & WikiHop \\
\midrule
Epochs & 20 & 10 & 10 & 10 \\
Prefix Len & 8 & 8 & 8 & 32 \\
Batch Size & \multicolumn{4}{c}{32} \\
Dropout & \multicolumn{4}{c}{0.1} \\
LR Warmup & \multicolumn{4}{c}{0.1} \\
LR Schedule & \multicolumn{4}{c}{linear} \\
Vocab size & \multicolumn{4}{c}{50265} \\
Loss & \multicolumn{4}{c}{cross-entropy} \\
Optimizer & \multicolumn{4}{c}{AdamW} \\
\bottomrule
\end{tabular}}
\caption{Hyperparameters used for our experiments. ``HY.'' and ``NG.'' denote the Hyperpartisan task and the 20-newsgroups tasks, respectively.}
\label{tab:hyperparameters}
\end{table}

\begin{figure}
    \centering
    \resizebox{\linewidth}{!}{
    \includegraphics{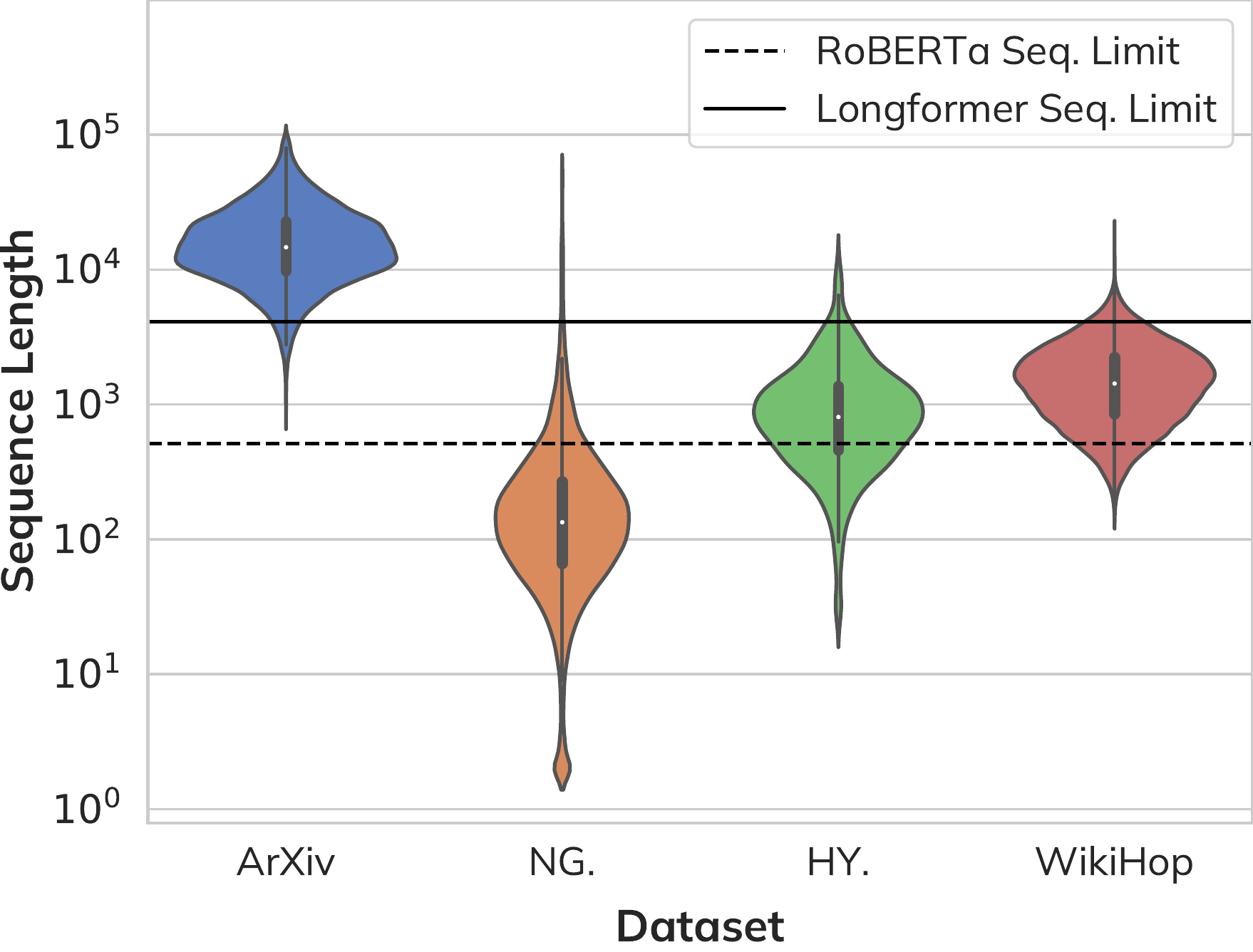}
    }
    \caption{Violin plot showing distribution of sequence lengths for each dataset. ``HY.'' and ``NG.'' denote the Hyperpartisan task and the 20-newsgroups tasks, respectively.}
    \label{fig:dataSizes}
\end{figure}

\paragraph{Training} For our experiments, we use and adapt the prefix-tuning implementation provided in \citet{liu-etal-2022-p}. Training was conducted on 12 NVIDIA GeForce 1080 Ti cards, for an estimated 2300 single GPU hours (including preliminary experiments). All models tested fit on a single card, so we did not use any model parallelism. Throughout experiments, we use gradient accumulation for an effective batch size of 32. We use early stopping for our hyperparameter search, and show results for the run with the best validation F1-score. For learning rate, we search between \{1e-2, 5e-2, 1e-3, 5e-3, 5e-4\} for prefix-based methods, and \{3e-5, 5e-5\} for fine-tuning. For kernelized prefix-propagation, we search for a scale factor (hyperparameter $\alpha$) of \{1e-2, 4e-2, 1e-3, 3e-3, 5e-3, 7e-3\} (after choosing the best learning-rate). Other hyperparameters are listed in Table~\ref{tab:hyperparameters}.

Despite seeding random number generators for Hugging Face's transformer library through the \code{set\_seed} method, slight deviations will propagate if using GPUs due to some non-deterministic CUDA methods that do not respect the seed setting mechanisms of Pytorch \citep{Torch:19}. Upon further analysis, we found this issue in non-deterministic algorithms to be widely overlooked in the field, and believe that this area needs further discussion in the research community. However, we note that our results should be reproducible when running across multiple seeds.

\paragraph{Task Details} All datasets used have a considerable portion of documents greater than \code{RoBERTa}'s max sequence limit of 512 tokens, as shown in Figure~\ref{fig:dataSizes}. Number of samples and number of classes for each dataset are in Table~\ref{tab:tasks}. 

For all classification tasks, we prepend a globally-attended \code{[CLS]} token to the start of the sequence and pass the output into a learned classification head. We truncate document lengths to 4096 and 512 tokens for \code{Longformer} and \code{RoBERTa}, respectively. For Hyperpartisan, we use the same data pre-processing and training split as \citet{Beltagy:20}. However, we noticed overlap between training and testing samples, so we instead show validation results. We use the ArXiv dataset from \citet{he-arxiv} that is available on Huggingface datasets (which we reviewed for correctness). The original dataset has labels leaked in the source text, so we use the \code{no\_ref} version that has those labels filtered. We use the 20-newsgroups and follow pre-processing as recommended by scikit-learn authors, removing headers, quotations, and signatures from each sample to prevent the model from learning spurious correlations.

\begin{table}
\centering
\resizebox{0.8\columnwidth}{!}{%
\begin{tabular}{c|ccc}
\toprule
Dataset & $n_{sample}$ & $n_{class}$ & $n_{train/dev/test}$ \\
\midrule
HY. & 645 & 2 & 80/10/10 \\
NG. & 18,846 & 20 & 60/20/20 \\
ArXiv & 33,388 & 11 & 85/7.5/7.5 \\
WikiHop & 48,867 & --- & 90/5/5 \\
\bottomrule
\end{tabular}}
\caption{Datasets used and their total size ($n_{sample}$), number of classes ($n_{class}$), and relative sizes of train, validation, and test splits ($n_{train/dev/test}$).}
\label{tab:tasks}
\end{table}

\begin{figure}
    \centering
    \resizebox{\linewidth}{!}{
    \includegraphics{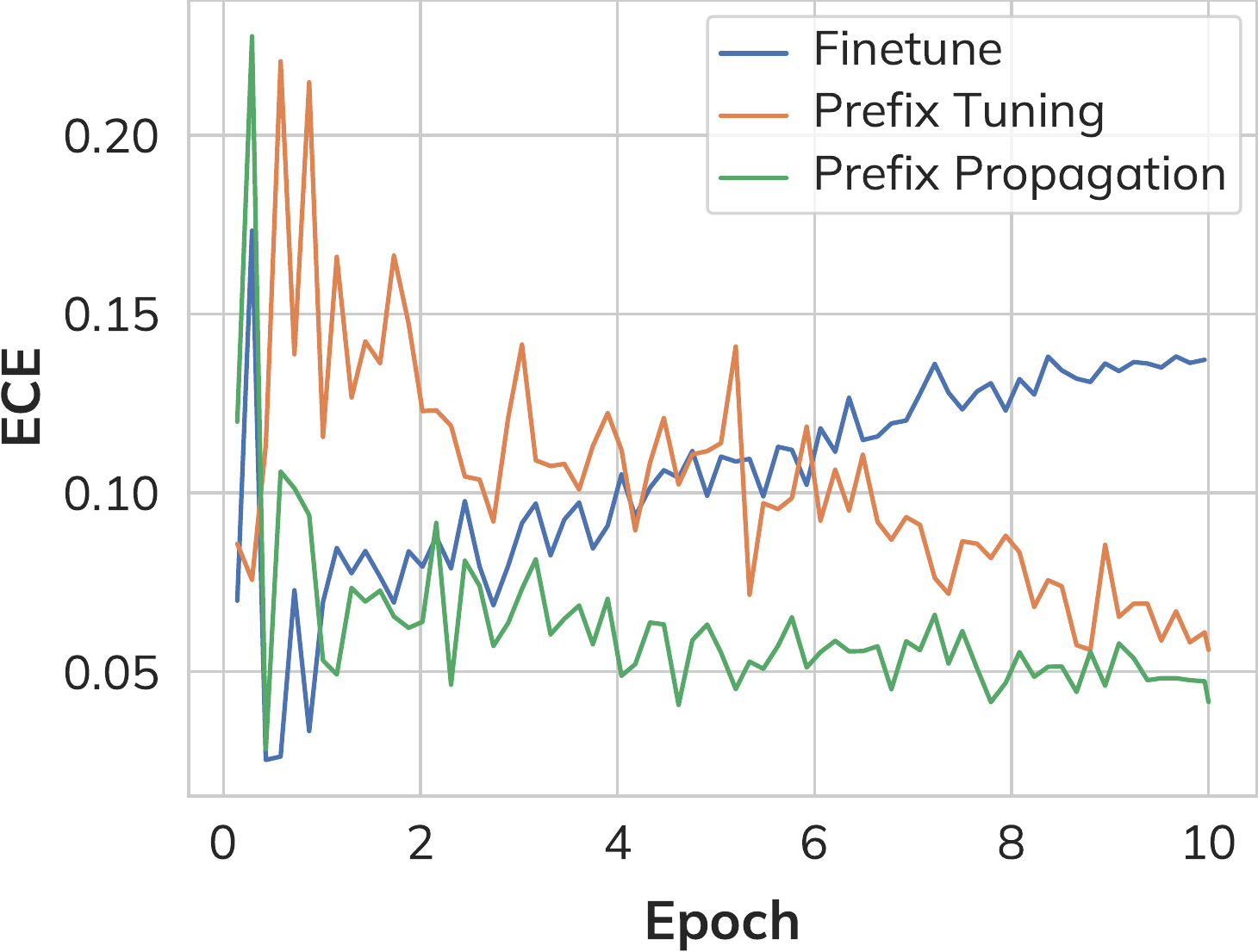}
    }
    \caption{Calibration (measured by ECE) of different tuning approaches using \code{Longformer-base} on ArXiv. Lower is better.}
    \label{fig:arxivCalibration}
\end{figure}

WikiHop instances include a question, candidate answers, and multiple context documents. For a fair comparison, we follow the WikiHop setup in \citet{Beltagy:20} to the best of our ability. In summary, we pass the dataset fields into the model in the format: \code{[q] <question> [/q] [ent] <candidate 1> [/ent] ... [ent]<candidate N> [/ent] [sep] <context 1> [sep] ... [sep] <context N>}. Because the context documents are often longer than the maximum sequence length of \code{Longformer}, we split the context documents into chunks of 4096 (or 512 for \code{RoBERTa}) and pass them separately through the model while concatenated to the question and candidate pair. We then train a classifier to predict a single logit for each \code{[ent]} token, take the average over all chunks, apply softmax, and finally use cross-entropy loss. We also train the new special tokens \code{[ent]} and \code{[q]} in prefix-based methods to better learn an effective representation (as they did not appear in pre-training).

\section{Impact of Training Time on ECE} \label{sec:ece_app}
Apparent in Figure~\ref{fig:arxivCalibration}, prefix-propagation is better-calibrated relative to other approaches throughout training. Prefix-tuning and fine-tuning however either start less calibrated or deviate from prefix-propagation as training progresses.

\section{Runtime Performance}
\begin{table}
\centering
\resizebox{\columnwidth}{!}{%
\begin{tabular}{c|cc}
\toprule
Method & Absolute Runtime (s) & Relative Runtime \\
\midrule
No PEFT & 2192 & 0\% \\
Prefix-Tuning & 2239 & +2.1\% \\
Prefix-Propagation & 2196 & +0.2\% \\
\bottomrule
\end{tabular}}
\caption{Runtime for inference using ``No PEFT'' (i.e., regular forward pass), prefix-tuning, and prefix-propagation. ``Relative Runtime'' is the runtime relative to ``No PEFT''.}
\label{tab:inferenceTime}
\end{table}

We test the inference time of the studied methods and show the results in Table~\ref{tab:inferenceTime}. We use the same 8000 randomly generated sequences of length 4096 across methods and test on a NVIDIA GTX 1080 Ti. We notice that prefix-propagation is slightly more efficient than prefix-tuning. We theorize that this discrepancy is caused by prefix-propagation only needing to concatenate a matrix in the first layer (and sum on the rest), whereas prefix-tuning concatenates before every layers.

\end{document}